%% file: tmi.tex
\pdfoutput=1
\documentclass[journal,twoside, print]{ieeecolor}
\usepackage{tmi}
\usepackage{cite}
\usepackage{amsmath,amssymb,amsfonts}
\usepackage{algorithmic}
\usepackage{graphicx}
\usepackage{textcomp}
\newcommand{\bfsection}[1]{\vspace*{0.1cm}\noindent\textbf{#1.}}

\def\BibTeX{{\rm B\kern-.05em{\sc i\kern-.025em b}\kern-.08em
    T\kern-.1667em\lower.7ex\hbox{E}\kern-.125emX}}
\begin{document}
% \lipsum[1] % filler text

%\title{Synapse Detection Challenge towards Connectomes of the Whole Microwasp Brain}
\title{An Out-of-Domain Synapse Detection Challenge for Microwasp Brain Connectomes}

\author{Jingpeng Wu, \and Yicong Li, \and Nishika Gupta, \and Kazunori Shinomiya, \and Pat Gunn, \and Alexey Polilov, \and Hanspeter Pfister, \and Dmitri Chklovskii, \and Donglai Wei
\thanks{J. Wu, K. Shinomiya, P. Gunn, and D. Chklovskii are with Flatiron Institute, New York, NY 10010 USA.}
\thanks{Y. Li and H. Pfister are with Harvard University, Cambridge, MA 02138 USA.}
\thanks{N. Gupta is with Birla Institute of Technology and Science-Pilani, Vidya Vihar, Pilani 333031, Rajasthan, India}
\thanks{A. Polilov is with Faculty of Biology, Lomonosov Moscow State University, Moscow, 119234 Russia.}
\thanks{D. Wei is with Boston College, Chestnut Hill, MA 02467 USA.}
}

\maketitle

\begin{abstract}
The size of image stacks in connectomics studies now reaches the terabyte and often petabyte scales with a great diversity of appearance across brain regions and samples. 
However, manual annotation of neural structures, \textit{e.g.,} synapses, is time-consuming, which leads to limited training data often smaller than 0.001\% of the test data in size.
Domain adaptation and generalization approaches were proposed to address similar issues for natural images, which were less evaluated on connectomics data due to a lack of out-of-domain benchmarks.
%Therefore, it is critical to design benchmark challenges to foster the development domain generalization methods .
%most existing connectomics benchmarks were constructed to have train and test data with similar appearance, 
%focus on fostering machine learning methods that excel on in-domain evaluation
%Extracting information from these images is constrained by time-consuming manual annotation. In contrast, machine learning models can be rapidly trained using a tiny portion of the larger dataset. In this approach, the generalization capability of the model is critical for the success of large-scale inference. 
%However, the challenge of targeting and optimizing this generalization capability in whole-brain connectomic studies has not been met. 
This challenge aims to push the boundary of out-of-domain generalization methods for large-scale connectomics applications. To facilitate this challenge, we annotated 14 image chunks from a biologically diverse set of \textit{Megaphragma viggianii} brain regions in three whole-brain datasets. Successful algorithms that emerge from our challenge could potentially revolutionize real-world connectomics research and further efforts that aim to unravel the complexity of brain structure and function.

%The size of image stacks in connectomic studies now reach the terabyte or even petabyte scale. Extracting information from these images is constrained by time-consuming manual annotation. In contrast, machine learning models can be rapidly trained using a tiny portion of the larger dataset. In this approach, the generalization capability of the model is critical for the success of large-scale inference. However, the challenge of targeting and optimizing this generalization capability in whole-brain connectomic studies has not been met. This challenge aims to push the boundary of generalization capability of machine learning models for real-world connectomics applications. To facilitate this challenge, we painstakingly annotated eighteen image chunks from a diverse set of \textit{Megaphragma viggianii} brain regions in three whole-brain datasets. Successful algorithms that emerge from our challenge can potentially revolutionize real-world connectomics research and further efforts that aim to unravel the complexity of brain functions.
\end{abstract}

\begin{IEEEkeywords}
Electron Microscopy, Synapse Detection, Machine Learning, Point Detection, Connectomics, Brain
\end{IEEEkeywords}

\section{Introduction}
\label{sec:introduction}
Neurons are the basic functional units of the brain and are connected by synapses. Synaptic connectivity constrains  information flow. Knowing synaptic connectivity is thus essential for understanding brain function and dysfunction.

Neurons can be long enough to span brain hemispheres and specifically connect to other neurons with nanometer-sized synapses. In order to reconstruct whole neurons, along with synapses, we need imaging methods with both a large field of view and nanometer resolution. The development of \textit{Volume Electron Microscopy} has met these requirements~\cite{peddie2014exploring,briggman2012volume,kornfeld2018progress,peddie2022volume, kleinfeld2011large}. As a result, many terabyte and petabyte-scale image volumes are being produced~\cite{microns2021functional, shapson2021human}. Manual annotation for all the structures in these datasets is impractical~\cite{motta2019big}. Techniques involving machine learning~\cite{plaza2014toward,chklovskii2010semi}, especially \textit{Deep Learning}~\cite{lecun2015deep}, can label such large-scale images automatically with good accuracy~\cite{lee2019convolutional}.

\begin{figure}
    \centering
    \includegraphics{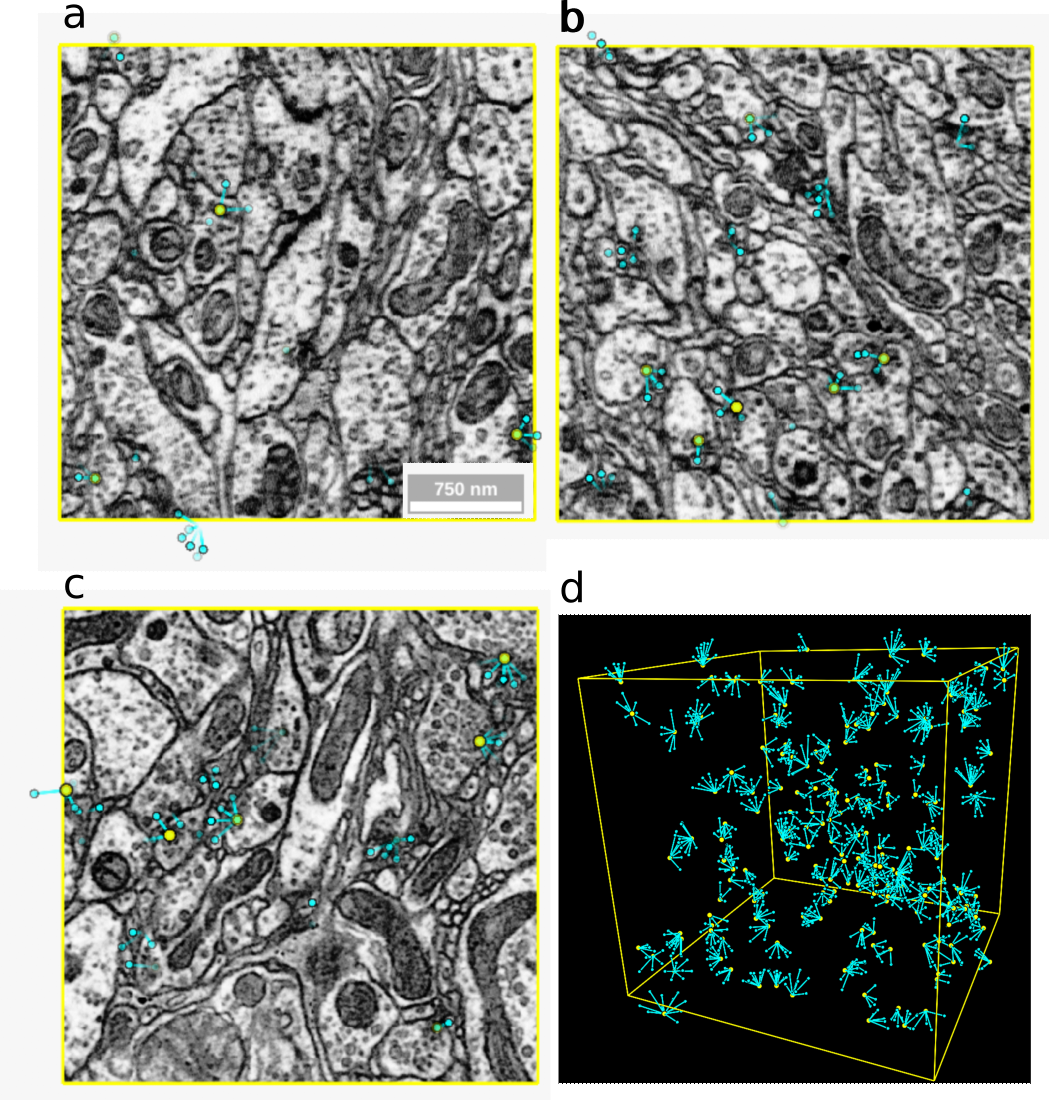}
    %\caption{Image chunk with manual annotation of point pairs. (a,b,c) The ZY, XZ, and XY planes are shown with manually annotated synapses. (d) 3D visualization of all the annotated synapses. Yellow dots represent presynapses and blue dots are postsynapses. The image chunk size is $416 \times 416 \times 416$ voxels and the physical size is $3.328~\mu m^3$.}
    \caption{Synapse detection from 3D electron microscopy (EM) image volume. (a,b,c) The ZY, XZ, and XY planes of the 3D volume with manually annotated synapses; (d) 3D point cloud visualization of the annotated synapses: presynapses represented as yellow dots and postsynapses as cyan dots and edges connected to the corresponding presynapses.}
    \label{fig:ground_truth_data}
\end{figure}

MICCAI Challenge on Circuit Reconstruction from Electron Microscopy Images (CREMI)~\footnote{https://cremi.org/} provided annotated ground truth data for training and performance evaluation. These efforts had already substantially facilitated computer vision research and helped the connectomics community to get more accurate automated neuron segmentation and synapse detection. However, CREMI still lacks coverage that we aim to address in this new challenge:

\begin{itemize}
    \item CREMI dataset images are acquired using Serial Section Transmission Electron Microscopy (SS-TEM) with anisotropic voxel size. One missing task is to develop the analysis of another advanced imaging method, Focused Ion Beam Scanning Electron Microscopy (FIB-SEM), with isotropic voxel size. The voxel size of our images is $8 \times 8 \times 8~nm$ compared with $4 \times 4 \times 40~nm$ in CREMI images. A detailed comparison of imaging methods is reviewed in~\cite{briggman2012volume}. 
    
    \item In real-world connectomic studies, models are normally trained using a small fraction of the terabyte or petabyte scale dataset since manual annotation is time-consuming and tedious. Thus, the ability to generalize is vital for real-world connectomic applications. Our challenge, therefore, focuses on testing generalization capability. In contrast with ground truth data from the same image stack where the test volumes are close to the training volumes in CREMI, we built ground truth volumes from three brain samples and provide a diverse set of test volumes. In total, we have annotated 14 ground truth volumes compared with 6 volumes in the CREMI challenge.
    
    \item There is a separate synaptic cleft detection task in CREMI. The synaptic cleft in our images is not clear due in part to smaller neurons and lower planar resolution. As a result, it is more challenging to detect post-synaptic regions in our images.
    
    \item For each synapse, CREMI labeled multiple point pairs across the synaptic cleft and there are many such points in each presynapse. Since there is no clear synaptic cleft in our images, but clear presynaptic motifs, also known as T-bar ribbons in the insect nervous system, in the bouton, we label one point in each T-bar. Thus, the distance from pre-synaptic points to corresponding post-synapses is much longer than the cross-membrane distance in CREMI. This requires a large field of view in the machine learning model.

    \item The mushroom body neurons in insects have distinct synapses compared to other neurons. No such volumes were present in the CREMI challenge.
    
    \item CREMI provides manual cell labeling for all the volumes that are used in the synapse detection evaluation. In contrast, we do not have manual labels for cells making them unavailable for training and quantitative evaluation.
    
    % \item Due to miniaturization in evolution~\cite{polilov2016size, polilov2012smallest}, the synapse density in our microwasp brain is much higher than fly and most mammalian brains. 
\end{itemize}

In short, an immense diversity of neurons and synapse textures exists and it is challenging to maintain consistent accuracy across regions in the whole brain. Here, we focus on testing the generalization capability of algorithms. We hope that successful algorithms emerging from this challenge would reduce the required amount of ground truth volumes in real-world connectomics projects. 

\input{data.tex}
\input{baseline.tex}
\input{plan.tex}

\section{Acknowledgement}
We appreciate the hard work of ground truth annotators: Diane Nguyen, Chi-Yip Ho, Aya Shinomiya, Sonia Villani, and Myisha Thasin.

The images were acquired and aligned by Song Pang and C. Shan Xu in the lab of Harald Hess at Janelia Research Campus. This is supported by the Howard Hughes Medical Institute.

The research at Flatiron Institute and the competition award are funded by Simons Foundation. The sample preparation is supported by Russian Science Foundation (project no. 22-14-00028 to AP).

This work was partially supported by NSF grant NCS-FO-2124179.

\bibliographystyle{IEEEtran}
\bibliography{tmi}

\newpage
\input{bio.tex}

\end{document}

%% file: data.tex
\section{Data and Challenge}
\subsection{Dataset Design}
\begin{figure}
    \centering
    \includegraphics{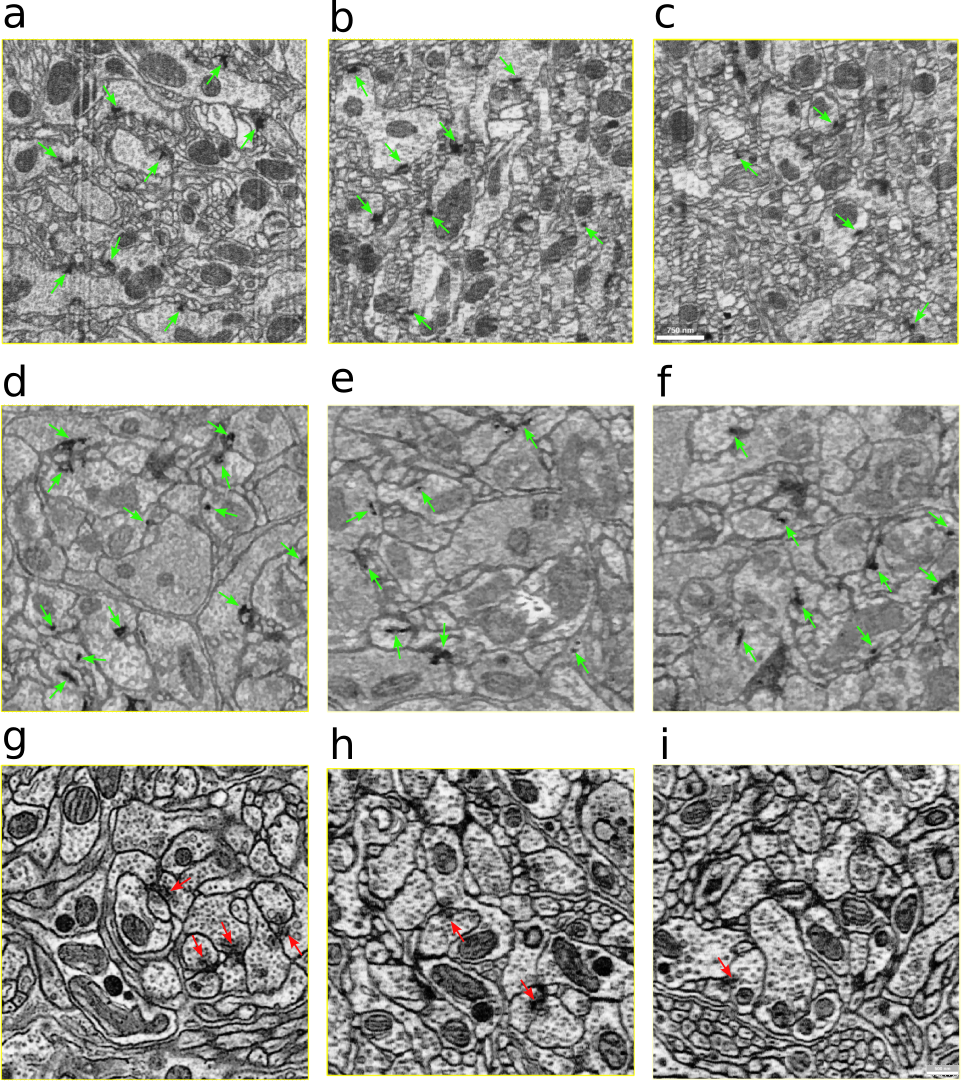}
    \caption{Images from three specimens. The columns of images are XY, XZ, and YZ planes from left to right. The images are from specimens one to three from top to bottom. The arrows indicate T-bars identified in the section. The red arrows indicate T-bars from mushroom bodies specifically.}
    \label{fig:images}
\end{figure}

\bfsection{Task description}
In the \textit{Megaphragma} brain, a chemical synapse consists of a presynaptic terminal, accompanied by an electron-dense motif called a T-bar, and multiple postsynaptic sites characterized by electron-dense regions (Fig.~\ref{fig:images}). 
%A T-bar is comprised of a platform and a pedestal that is attached to the plasma membrane at a synaptic active zone. T-bars are indicated in the Figure~\ref{fig:images}.
We define two computational tasks. (1) Presynaptic T-bar detection: predict the center location of presynaptic T-bar structure from input image volumes. (2) Postsynaptic site detection: predict the postsynaptic site locations given the presynaptic T-bar locations and the input image volumes.

\bfsection{Data acquisition}
We focus on \textit{Megaphragma viggianii} because it has both a small brain size and complex behaviour. These wasps have evolved anucleate neurons, likely due to the selective pressure that has driven miniaturization ~\cite{polilov2012smallest}. The scientific significance is detailed in previous publications~\cite{makarova2021small, polilov2012smallest, polilov2016size}. 
The whole head of \textit{Megaphragma} was stained with heavy metal and embedded in resin~\cite{polilov2021protocol}. Subsequently, the sample was imaged using enhanced Focused Ion Beam Scanning Electron Microscope (FIB-SEM)~\cite{Knott2959, xu2017enhanced, xu2020transforming} with an isotropic voxel size of $8 \times 8 \times 8~nm$. 

We make the following design choices:
\begin{itemize}
    \item Cross-sample variation. We imaged three brain specimens that are illustrated in Figure~\ref{fig:images}.
    We densely annotated 14 image chunks, each of which has 400 $\times$ 400 $\times$ 400 voxels: three chunks are from specimen one, three chunks are from specimen two, and eight chunks are from specimen three. 
    \item Cross-region variation. Different brain regions include MB, AL, PLP, GNG, CBL, etc. In order to challenge the generalization capability of machine learning models, we provide annotations for one volume per region above for the third specimen. % , GNG, Protocerebrum, MB-ML, and MB-VL
    \item Challenging cases. In the mushroom body, multiple Kenyon cell terminals connect to an output neuron terminal, making a rosette-like structure. Presynaptic terminals of Kenyon cells in a rosette lack platforms and are smaller than typical T-bars.
    \item Dataset split. In summary, we will use five volumes for training and validation, and nine volumes for testing. To evaluate the out-of-domain performance of models, we split the eight annotated volumes in the third specimen into 5/3 for train/test. Users can split the training set to some validation set themselves. We use the three volumes in the first and second specimens as challenge tests.
\end{itemize}

\bfsection{Data annotation}
We used CATMAID~\cite{saalfeld2009catmaid} and NeuTu~\cite{zhao2018neutu} with DVID~\cite{katz2019dvid} to annotate the presynapses and postsynapses using point pairs. We annotated each T-bar with a point and connected it with PSDs - PostSynaptic Densities. Figure~\ref{fig:ground_truth_data} illustrates the point annotation results in an image chunk. 
In synapse annotation, each T-bar is associated with corresponding postsynaptic terminals to represent a synaptic connection (Fig.~\ref{fig:ground_truth_data}). A T-bar glyph should be placed at the connecting point of the platform and the pedestal. A large platform may have contacts with more than one pedestal, in which case each contact point should be annotated as a separate T-bar. Profiles are annotated as postsynapses if PSDs are clearly visible. If PSDs are not recognizable due to image quality, all bodies within $40 nm$ from the edge of the platform are considered to have PSDs.

\begin{figure}
    \centering
    \includegraphics{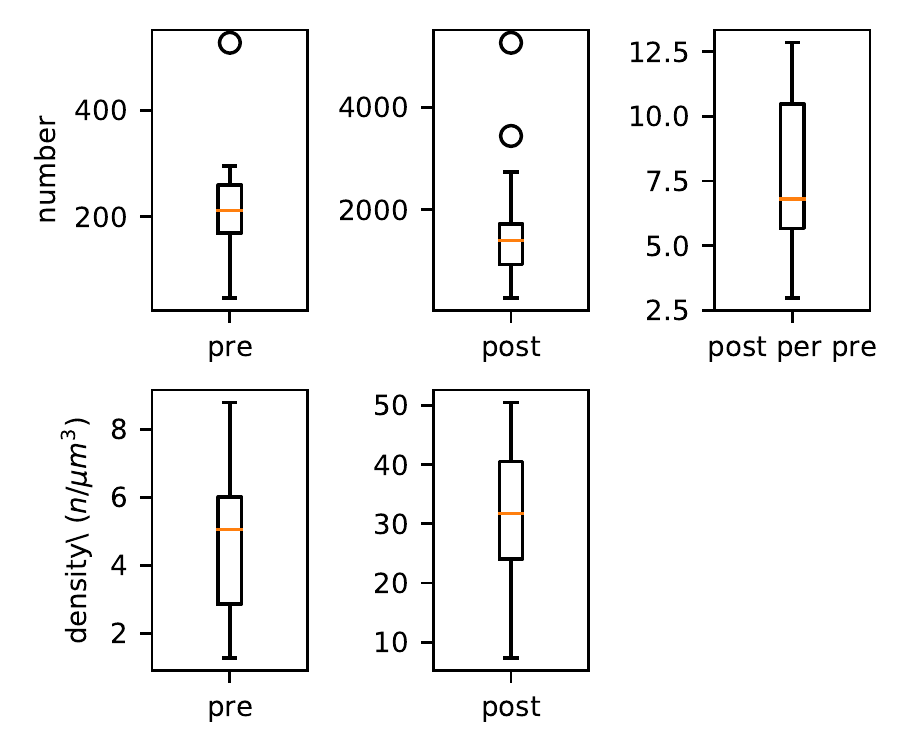}
    \caption{Statistics of ground truth in each volume.}
    \label{fig:statistics}
\end{figure}

\bfsection{Data statistics}
As illustrated in Figure~\ref{fig:statistics}, we performed some basic statistical analysis in all the ground truth volumes. There are about 225±103 T-bars and about 1735±1253 PSDs in each volume. %\textcolor{red}{Note that the volume size varies from .} 
There are about 7±3 PSDs connecting to each T-bar. The T-bar density is about 4.7±2.1 per $\mu m^3$ and the PSD density is about 31.2±12.0 per $\mu m^3$.  

%\red{table: distribution of chunks. sample number: brain region training, validation, public leaderboard test, private leaderboard test}

\subsection{Challenge Design}

\bfsection{Submission method}
For submission, participants should create a .zip file that includes dense predictions for each task. The write-up should also include an algorithm speed assessment following the template released by the committee and should state if any public datasets were used (for pre-training, transfer learning, etc.). No code is required for submission.

\bfsection{Life cycle type} The challenge will continue to accept submissions after the deadline for continuous benchmarking. Results submitted after the deadline will not be included in the prize competition and the publication of the challenge.

\bfsection{Publication policy} The three top-performing teams are eligible to participate in a joint publication with the committee submitting to IEEE Transactions on Medical Imaging (TMI). There is a fixed maximum of two authors per team. The committee may also invite teams that submit particularly novel solutions to join as co-authors.

\bfsection{Organizer participation policy} Committee members will not participate in the challenge but only provide baseline results.
%are not be listed on the leaderboard. Those that were involved in annotating the data are not eligible to participate in the challenge.

\bfsection{Award policy}
A certificate will be awarded to challenge top-3 teams (1 winner and 2 runner-ups). Three iPads with different configurations will be awarded to the top-3 teams. 
% We are also applying for some funding support from the Simons Foundation.

\bfsection{Accessibility} Because the committee will only be releasing some small image chunks, there is no need for external computing resources.

% \begin{table}
% \caption{Ground Truth Volumes}
% \label{table: ground truth volumes}
% \begin{tabularx}{\textwidth}{@{} l *{3}{C} c @{}}
% \toprule
% volume ID  & Brain Region & Set \\ 
% \midrule
% SA_1    & Left {OL}      & Train\\ 
% SA_2    & Left {VLP}     & Train  \\ 
% \bottomrule
% \end{tabularx}
% \end{table}

% \begin{table}[htb]
% \caption{Ground Truth Volumes}
% \label{table: ground truth volumes}
% % \resizebox{\columnwidth}{!}{%
% \begin{tabular}{c*{3}}
% % \toprule
%     & \text{Volume ID}  & \text{Brain Region} & \text{Set} \\
% % \midrule
% {SA_1}    & Left {OL}      & Train \\ 
% {SA_2}    & Left {VLP}     & Train  \\ 
% % \bottomrule
% \end{tabular}
% % }
% \end{table}
% SA = Sample 3, SB = Sample 2, SC = Sample 1

% \begin{table*}[h]
%   \centering
%   \caption{Comparison table}
%   \label{tab2}
%   \begin{tabular}{18* c}
%     \toprule
%     \multicolumn{1}{}{}Metric & \cite{dagher2018ancile} & \cite{tripathi2020sms} & %
%       \cite{zheng2018blockchain} & \cite{gordon2018blockchain} & Our architecture\\
%     \midrule
%     User Centric                        & Y & N & Y & Y & Y\\
%     User Authentication                 & N & N & N & N & Y \\
%     Privacy of data owner               & Y & Y & Y & Y & Y\\
%     Store personal data into blockchain & N & N & N & N & Y\\
%     Transparent policy                  & N & N & N & N & Y\\
%     Use of cryptographic functions      & N & Y & Y & N & Y\\
%     Blockchain based                    & Y & Y & Y & Y & Y\\
%     \bottomrule
%   \end{tabular}
% \end{table*}

%% file: baseline.tex
\begin{figure*}
    \centering
    \includegraphics[width=0.8\linewidth]{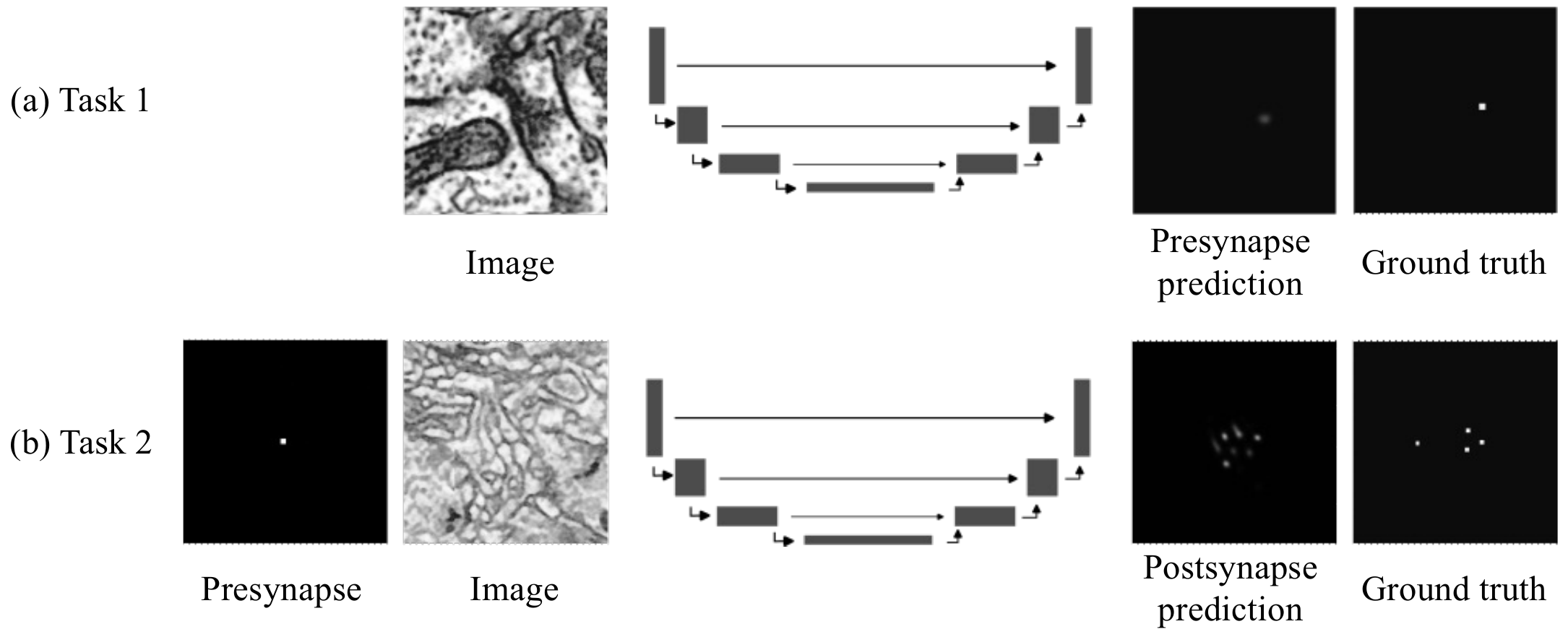}
    \caption{Baseline method using 3D U-Net. (a) Training of T-bar detection network. The patches are randomly sampled so that some patches might not contain any T-bar. (b) Training of post-synapse detection network. The T-bar is in the center of each patch and a fixed patch with a central point is used as a channel of input patch. Note that the illustration is 2D while both the patches and network are 3D and of isotropic size.}
    \label{fig:baseline}
\end{figure*}

\section{Baseline and Evaluation}
\subsection{Baseline Method}
\bfsection{Training} As illustrated in Figure~\ref{fig:baseline}, we built a baseline method using 3D U-Net~\cite{ronneberger2015u, cciccek20163d}. For T-bar detection training, random 3D image patches are sampled, augmented and fed into the network. The annotated points are then transformed into voxel cubes with a size of three voxels. For postsynaptic detection, a presynapse is sampled from the ground truth dataset and the image patch around the presynapse cropped out and used as input for network. The network then predicts the postsynaptic probability map directly. The 3D U-Net architecture was modified from a previous synapse detection method~\cite{turner2020synaptic}. All kernel sizes are changed to isotropic. 

\bfsection{Predicting presynaptic T-bars} For the inference of T-bars, we scan the image volumes using a 3D sliding window using chunkflow~\cite{wu2021chunkflow}. The window patches overlap with each other by 50\% yielding eight fold coverage of each voxel. The overlapping network output patches are blended together to produce a probability map. To find T-bar locations, we detect local maxima in the probability map. 

\bfsection{Predicting postsynaptic sites} For the inference of postsynapses, we scan all the detected T-bars, extract the surrounding image patch, and perform the network inference to produce the postsynapse probability map using chunkflow~\cite{wu2021chunkflow}. Finally, we detect postsynapse points using the same algorithm to detect local maxima.

% \vspace{-0.25cm}
\subsection{Evaluation Metrics}

Participants will be required to submit their results, detection of presynaptic T-bar and postsynpatic sites including their connectivity, for evaluation. The detection accuracy of a submission will be evaluated by first solving an assignment problem minimizing the Euclidean distance between detected synapses and ground truth synapses to find true matches, and then calculating the F1-score. Formally, given a set of detected synapses ($D$) by participants and a set of ground truth synapses ($G$), we want to find a bijection $f: D \rightarrow G$ to minimize the following cost function:
\begin{equation}
    \sum_{d \in D} C (d, f(d)),
\end{equation}
where $C(\cdot)$ denotes the Euclidean distance of a matched pair. Next, F1-score is defined as:
\begin{equation}
    F_1 = \frac{2TP}{2TP+FP+FN},
\end{equation}
where TP is the true positive, FP is the false positive, and FN is the false negative.

\bfsection{Task 1: Presynaptic T-bar detection} Detected T-bars are considered to be potential matches to the ground truth T-bars. After solving the assignment problem, an unmatched detected T-bar will be counted as one FP, an unmatched ground truth T-bar will be counted as one FN, and a falsely matched presynapse pair as one FP and one FN. The presynapse detection accuracy will be expressed as the F1-score calculated using TPs, FPs, and FNs. Notably, we use a threshold to determine the falsely matched pairs. If the distance between the detected T-bar and the matched ground truth T-bar in a pair exceeds the threshold, this pair will be considered as a false match.

%\textbf{The second step is to evaluate the detection accuracy of the postsynapses.}
\bfsection{Task 2: Postsynaptic sites detection}
Since our data involves one-to-many synapse connections, for each matched T-bar pair, we compare its connected postsynapses by solving the assignment problem mentioned above. An unmatched but detected postsynapse will be counted as one FP, an unmatched ground truth postsynapse will be counted as one FN, and a falsely matched postsynapse pair will be counted as one FP and one FN. The F1-score for postsynapse detection can be computed using TPs, FPs, and FNs.

\bfsection{Ranking mechanism} To determine each participant's position on the leaderboard, we will use the arithmetic mean of F1-scores of Task 1 and Task 2 as the overall score.

%% file: plan.tex
\section{Plan and Schedule}
\bfsection{Schedule} The committee has set the following dates: 
\begin{itemize}
    \item 2023.01.31: Website launch on https://codalab.org/; release of training and test data in H5 format; release of evaluation code, example submission, and utility functions
    \item 2022.01.31-2023.03.31: Registration and submission period
    \item 2023.03.31: Leaderboard release and invitation for workshop manuscript submission
    \item 2023.04.18: Presentations at the workshop of ISBI 2023  
\end{itemize}
%training data and code for metric computation, single or multiple phases of test data, format and dummy examples of the submission files, format of the participant workshop manuscripts, submission of test results, submission of the manuscripts, declaration of the leaderboard, presentations at the workshop, and post-workshop leaderboard release

%\bfsection{Data to be ready}

%\bfsection{Source code release} Participants are encouraged but not required to release source code. 

\bfsection{Estimated number of participants} 
We expect between 10-20 teams to participate for the following reasons:

% \begin{itemize}
    % \item 3D instance segmentation is an established paper submission track for major biomedical image analysis conferences. For MICCAI 2020 alone, there were more than 50 accepted papers. We believe that our challenge can attract a broader audience with our large-scale saturated 3D instance segmentation datasets using different training supervision strategies.
% \end{itemize}

\begin{itemize}
\item The prior MICCAI 2015 CREMI challenge on synapse detection drew more than 10 teams. Thus, we expect at least a similar turnout for our challenge. 
\item 3D object detection is an established paper submission track for major biomedical image analysis conferences. For MICCAI 2022 alone, there were more than 25 accepted papers~\cite{mccai}. This path toward authorship will likely catch the attention of prospective participants that may otherwise not be interested.
\item In addition, as volumes in our dataset have big domain gaps, we expect general machine learning researchers in domain adaptation will be interested to develop novel methods through our challenge.
\item Dr. Wei, one of our organizers, led the ISBI 2022 MitoEM challenge for 3D instance segmentation that drew 20 participants. He will reach out to the microscopy image segmentation and detection community for our challenge.
\end{itemize}

%% file: bio.tex
\section{Biosketch}

Dr. Jingpeng Wu is an Associate Research Scientist at the Flatiron Institute. He is working on mapping neurons based on high-resolution Electron Microscopy images. Prior to joining Flatiron, he was an Associate Research Scholar at Princeton University, where he worked on petabyte-scale neuron reconstruction based on electron microscopy images using deep learning and cloud computing technologies. He has a Ph.D. in Biomedical Engineering from Huazhong University of Science and Technology in China. During his Ph.D., he worked on large-scale neuron and blood vessel tracing based on light microscopy images of whole mouse brains.\\

Mr. Yicong Li is a Ph.D. student in computer science at Harvard University, advised by Prof. Hanspeter Pfister. He is working on biomedical image analysis, AI for healthcare,  computer vision, and connectomics. His research has been published in top venues including CVPR, MICCAI, etc. He obtained his master's degree at Tsinghua University and his bachelor's degree at Sichuan University.\\

Ms. Nishika Gupta is a final-year undergraduate student currently pursuing a double major in Mathematics and Computer Science at BITS Pilani. She is expected to graduate in May 2023. Currently, she is working on connectomics under the guidance of Prof. Donglai Wei, which aims to solve the comprehensive connectivity graphs in animal brains to shed light on the underlying mechanism of intelligence and inspire better treatment for neurological diseases. Throughout her undergraduate studies, she has been actively involved in research projects in computer vision and is always keen to explore and develop real world applications of AI.\\

Dr. Kazunori Shinomiya is a Research Scientist at Flatiron Institute. He is working on the brain connectome project of the miniature wasp, \textit{Megaphragma viggianii}. He is responsible for reconstruction, annotation, and analysis of neuronal circuits. Before joining Flatiron, he developed connectomics of the fruit fly brain using electron microscopy as part of the FlyEM Project Team at HHMI Janelia Research Campus. He earned a PhD in Computational Biology from the University of Tokyo, Japan, where he helped to develop a standard atlas and nomenclature system for the insect brain using confocal laser scanning microscopy. He worked as a postdoctoral researcher at Dalhousie University, Canada, on the neuroanatomy of the fly visual system.\\

Pat Gunn is a member of the Scientific Computing Core at the Flatiron Institute. Prior to this, he worked at Dropbox, MongoDB and Spotflux. He also spent a decade as a member of the research staff at Carnegie Mellon University in Pittsburgh, where he successively worked in research groups focused on high-dimensional clustering, neuroscience and clustered operating systems. \\

Prof. Hanspeter Pfister is An Wang Professor of Computer Science in the School of Engineering and Applied Sciences at Harvard University. Before joining Harvard he worked for over a decade at Mitsubishi Electric Research Laboratories where he was Associate Director and a Senior Research Scientist. Dr. Pfister has a Ph.D. in Computer Science from the State University of New York at Stony Brook and an M.S. in Electrical Engineering from ETH Zurich, Switzerland. \textbf{He was also a co-organizer for ISBI 2021 MitoEM challenge.}\\

Prof. Alexey Polilov is head of the Department of Entomology, Faculty of Biology, Lomonosov Moscow State University. The main focus of his research is the functional and evolutionary morphology of the smallest insects. He pioneered the study of miniaturization in insects and was the first to study the external morphology and anatomy of many of the smallest insects of different orders. His work revealed unique structural features associated with extremely small body size. He obtained the degrees of Candidate of Sciences (PhD) and Doctor of Sciences (habilitation) at Lomonosov Moscow State University. His current research examines the effects of miniaturization in insects and encompasses not only morphology but also connectomics, genomics, the study of flight, and the study of cognitive abilities in very small insects.\\

Prof. Chklovskii is a Group Leader at the Flatiron Institute. He is also on the faculty of the NYU Medical Center. Before that, he was a Group Leader at the Janelia Research Campus where he initiated and led a collaborative project that assembled the largest connectome at the time, producing a comprehensive map of neural connections in the brain. Prior to that, he was an associate professor at Cold Spring Harbor Laboratory in New York, Sloan Fellow at the Salk Institute and Junior Fellow of the Harvard Society of Fellows. He holds a Ph.D. in physics from the Massachusetts Institute of Technology.\\

Prof. Donglai Wei is an assistant professor in the Computer Science Department at Boston College. His research focuses on developing novel registration and reconstruction algorithms for large-scale (currently petabyte-scale) connectomics datasets to empower neuroscience discoveries. During his Ph.D. at MIT under Prof. William Freeman, he worked on video understanding problems, including arrow of time and the Vimeo-90K benchmark. Since his postdoc at Harvard University, he has sought to reconstruct the brain's wiring diagram. \textbf{He was the lead organizer for the ISBI 2021 MitoEM challenge and co-organizer of the MICCAI 2020 RibFrac Challenge.}\\